\documentclass[10pt, a4paper]{article}

\usepackage[final]{lrec2026} 

\usepackage{comment}
\usepackage{cleveref}
\usepackage{booktabs}

\usepackage{array}
\usepackage{longtable}

\usepackage[textsize=tiny]{todonotes}
\usepackage{caption}

\title{Unrequited Emotions: Investigating the Gaps in Motivation and Practice in Speech Emotion Recognition Research}

\name{Taryn Wong\textsuperscript{1}, Zeerak Talat\textsuperscript{2}, Hanan Aldarmaki\textsuperscript{3}, Anjalie Field\textsuperscript{1}} 

\address{\textsuperscript{1}Johns Hopkins University, \textsuperscript{2}University of Edinburgh, \textsuperscript{3}MBZUAI \\
         	z@zeerak.org, hanan.aldarmaki@mbzuai.ac.ae, anjalief@jhu.edu\\}

\abstract{
Critical analyses of emotion recognition technology have raised ethical concerns around task validity and potential downstream impacts, urging researchers to ensure alignment between their stated motivations and practice.
However, these discussions have not adequately influenced or drawn from research on speech emotion recognition (SER). 
We address this gap by conducting a systematic survey of SER research  to uncover what stated motivations drive this work and if they align with the datasets and emotions studied.
We find that while SER research identifies appealing goals---such as well-situated voice-activated systems or healthcare applications---commonly-used datasets do not reflect these proposed deployment contexts, thus presenting a gap between motivations and research practices.
We argue that such gaps engender ethical concerns, and that SER research should reassert itself with concrete use-cases to prevent misinterpretations, misuse, and downstream harms.
 \\ \newline  }

\begin{document}

\maketitleabstract

\section{Introduction}

Emotion AI, including the detection of human emotions from video, image, audio, or text data using machine learning (ML) methods, has become a popular research area with increasing commercialization in a broad range of applications \cite{mcstay2018emotional}.
The growing adoption of this technology has raised concerns around its development and use. 
Researchers have questioned the overall task validity, e.g., the premise of using someone's outer appearance to infer inner character or state, particularly given the lack of consensus on how to define emotions among psychologists \cite{mcstay2019emotional,Stark2021
}.
Empirical qualitative studies have demonstrated that people view emotion AI as invasive and privacy-violating \cite{Andalibi2020,pyle2024us}, and its use in contexts like workplaces or hiring processes create emotional labor with little perceived benefit for those subjected to it \cite{Roemmich2023,pyle2024us}. 
Importantly, concerns raised are often specific to emotion AI, rather than technology more generally \cite{pyle2024us}.

To date, this work has had limited engagement with the abundance of research on Speech Emotion Recognition (SER). 
Research on task validity \cite{jacobs2021measurement} and ethics have focused on facial recognition or broadly construed AI systems where underlying data and models are unspecified \cite{Domnich2021ResponsibleAG,pyle2024us,Roemmich2023,karizat2024patent}.
While some studies have questioned the ethics and validity of SER \citep{katirai2023ethical,hartmann2024healthy,morozov2014save,testa2023privacy, juslin2005vocal}, no prior work has conducted a systematic investigation of SER research practices.

In this work, we follow the suggestion proposed by \citet{Stark2021} that ``[d]esigners and developers should think twice before embarking on emotion AI projects'' and we investigate to what extent SER research meets their proposed  ``necessary though not sufficient condition'' that projects have ``clear alignment between conceptual models, data, norms, and aims''. 
We formalize our study into three research questions: (1) What do SER researchers state as their motivations? (2) What purposes do popular datasets support? (3) Do datasets align with stated motivations?

RQ1 focuses on \textit{why} researchers work on SER, while RQ2 focuses on \textit{how}. 
RQ3 assesses the alignment between the \textit{why} and \textit{how}. 
We investigate these questions by conducting a systematic survey of 88 SER research papers, which we manually coded for stated motivations, specific emotions studied, and datasets used. 
We ultimately find a substantial gap between stated motivations and experimental setups, and we propose ways to reduce this gap by seeking suitable datasets or pursuing different motivations. Our work encourages future researchers to ``think twice'' about SER projects, and to seek alignment between objectives, research practices, and potential implications, rather than viewing SER as an isolated research task.\looseness=-1

\section{Methodology}
Our work uses similar systemic survey methodology as previous studies that reflect on practices in speech processing research and related disciplines \cite{blodgett2020language,field2021survey,birhane2022values,raff2024machine}. First, we queried Semantic Scholar for papers containing search terms ``speech'' and ``emotion'' and \{``recognition'', ``classification'', ``data'', or ``database''\} in their title or abstract, collecting a total of 7,486 after filtering out papers without publication venue information. 
Of the initial sample, 
we kept only those published in popular archival speech, natural language processing, or ML venues, which we determined by manually reviewing venues with the most papers in the initial sample,\footnote{ICASSP; ASRU; SLT; Interspeech; NeurIPS; ICLR; ICML; AAAI; IEEE/ACM Transactions on Audio, Speech, and Language Processing; IEEE Open Journal of Signal Processing; IEEE Journal of Selected Topics in Signal Processing; Computer Speech \& Language; Speech Communication; IEEE Transactions on Affective Computing; JMLR; ACL} resulting in 959 papers. 

We then randomly sampled 108 (stratified by publication year and venue) for in-depth manual analysis.

We used a standard inductive coding procedure to label the sampled 108 papers. 
First, three authors independently coded a set of 10 papers each (30 papers total), where each author first determined if the paper consisted of SER. 
Then, each author labeled the paper's stated motivations, specific emotions investigated, and data sets used in experiments or analysis.\footnote{We also initially coded acknowledged funding sources, but found they were too sparse to analyze in aggregate.} In labeling stated motivations, annotators also quoted the exact text from the paper describing these motivations. 
The annotators then discussed the labeling scheme and finalized a set of frequently-stated motivations for which to code. 
Annotators then re-coded the initial 30 documents under the revised scheme, and then each independently coded an additional set of 10 documents to ensure mutual understanding of the coding scheme. 
The remainder of the data was then singly coded. 
Finally, the annotators re-convened and discussed further revisions to the scheme, specifically adding categories of motivations that were found in the larger data sample and re-coded the data under the revised scheme.
Throughout this process, we removed 20 papers whose primary focus was not SER (e.g., papers that focused on depression detection, speech generation, speech representation, etc.),
leaving a final set of 88 papers.
The final set primarily consists of papers published in speech venues (Interspeech: 50 papers, ICASSP: 18, ASRU: 1, SLT: 2, Speech Communication: 4), as well as 9 papers from IEEE Transactions on Affective Computing. 
The final coding for all 88 papers is in \Cref{tab:allpapers} in the Appendix.

\section{Results}

\paragraph{What do SER researchers state as their motivations?}

\begin{table}
    \centering
    \scalebox{0.8}{
    \begin{tabular}{lc}
    \hline
Responsive bots: other HCI systems & 42.05\% \\
Responsive bots: voice assistants & 12.50\% \\
Responsive bots: car voice assistants & 6.82\% \\
\hline
Healthcare (mental health) & 18.18\% \\
Call screening & 17.05\% \\
Video games, toys, entertainment & 13.64\% \\
Education & 9.09\% \\
Paralinguistics / behavioral studies & 6.82\% \\
Social companion bots & 4.55\% \\
Lie detection & 3.41\% \\
\hline
Prior work & 27.27\% \\
Other & 14.77\% \\
    \end{tabular}
    }
    \caption{Percent of manually coded papers that reference each motivation. Motivations are grouped thematically. Percents do not add to 100 as papers often reference multiple motivations.}
    \label{tab:motivations}
\end{table}

\Cref{tab:motivations} presents the 12 stated motivations identified in our coding scheme with examples for each label in \Cref{sec:app_examples}.
The most common motivation was enabling human-computer interaction systems to respond more naturally and effectively (``Responsive bots''), which we subdivide into ``voice assistants'', ``car voice assistants'', and ``other HCI systems'' (typically unspecified systems).
 Other common motivations focused on specific deployment contexts, including healthcare (almost always mental health), 911 or customer service calls, education, and entertainment. Less common motivations 
 included social companion bots and studies of human behavior or paralinguistics. 

 While most studies include motivations drawn from prior SER work, in data coding, we use the label ``Prior work'' to indicate that the paper drew motivation \textit{exclusively} from prior SER research without without specifying how they expect this technology to be used.
  We also identified some papers with ``Other'' motivations. 
 In practice, this label typically indicates that the paper includes a long list of applications, without depth of focus on any one. 
\begin{figure}[h]
    \centering
    \includegraphics[width=\linewidth]{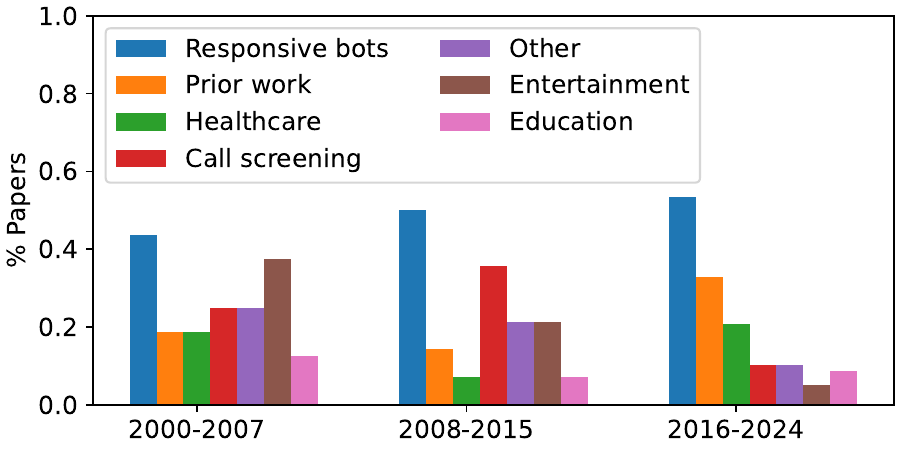}
    \caption{Percent of papers in each time window that reference each motivation. The number of papers in each time window are [16, 14, 58] respectively.  We drop infrequent motivations for readability.}
    \label{fig:motivations_overtime}
\end{figure}
In \Cref{fig:motivations_overtime} we additionally report shifts over time. While ``Responsive bots'' has consistently been a motivating application, ``Call screening'' and ``Entertainment'' have declined from 2000-2015 papers to 2016-2024 papers in our sample.

 \begin{table}
    \centering
    \scriptsize
    \begin{tabular}{p{0.15\linewidth} | p{0.13\linewidth} | p{0.18\linewidth} | p{0.35\linewidth}}
\toprule
\textbf{Dataset}  & \textbf{\% Papers} & \textbf{Acted/Spont.} & \textbf{Emotions} \\
\hline
IEMOCAP  & 40.91\% &  Acted (naturalistic)
& \textcolor{violet}{Angry}, Neutral, \textcolor{violet}{Sad}, \textcolor{violet}{Happy}, \textcolor{violet}{Fearful}, \textcolor{violet}{Surprised}, Frustrated, Excited, \textcolor{teal}{Valence/Sentiment}, \textcolor{teal}{Arousal/Activation}, \textcolor{teal}{Dominance}\\
\hline
EMO-DB &  17.05\% & Acted & \textcolor{violet}{Angry}, Neutral, \textcolor{violet}{Happy}, \textcolor{violet}{Fearful}, \textcolor{violet}{Disgusted}, Anxious, \textcolor{violet}{Bored} \\
\hline
RAVDESS & 9.09\% & Acted &Produced at both normal \& strong intensities: \textcolor{violet}{Angry}, Neutral,\textcolor{violet}{Sad}, \textcolor{violet}{Happy}, \textcolor{violet}{Fearful}, \textcolor{violet}{Surprised}, \textcolor{violet}{Disgusted}, Calm\\
\hline
SUSAS &  6.82\% & Both & Stressed\\
\hline
MSP-Improv & 6.82\% & Acted (naturalistic) & \textcolor{violet}{Angry}, Neutral, \textcolor{violet}{Sad}, \textcolor{violet}{Happy} \\
\hline
RECOLA &  6.82\% & Spontaneous & 
\textcolor{teal}{Valence/Sentiment}, \textcolor{teal}{Arousal/Activation},

Social behaviors rated as either positive or negative: Agreement, Dominance, Engagement, Performance, Rapport\\
\bottomrule
\textbf{Exclusive} & 35.23\% & & Paper exclusively used data in the top 6 \\
\textbf{Custom} & 22.73\%  & & Paper custom-collected data \\
\textbf{Other} & 45.45\%  & & Paper used data not in the top 6\\
\hline
    \end{tabular}
    \caption{Datasets used in our analysis corpus $\ge5$ times. We provide mode of collection and emotions labels for each dataset for reference. Categorical labels from Basic Emotion Theory are in \textcolor{violet}{violet}. Axes from dimensional theories are in \textcolor{teal}{teal}.}
\label{tab:datasets}
\end{table}

\paragraph{What purposes do popular datasets support? }
\label{results_rq2}

In \Cref{tab:datasets}, we identify ``popular'' datasets as ones used by at least five papers: IEMOCAP \citep{busso2008iemocap}, EMO-DB \citep{emodb}, RAVDESS \citep{livingstone2018ryerson}, SUSAS \citep{hansen1997getting}, MSP-Improv \citep{Busso_2017}, and RECOLA \citep{ringeval2013introducing}. These six datasets support the majority of research in our corpus: 63.64\% of papers use at least one of them, and 35.23\% exclusively use them. 
Of the six popular datasets, four (IEMOCAP, EMO-DB, RAVDESS, MSP-Improv) consist of acted emotions and two (SUSAS, RECOLA) consist of emotions captured in spontaneous speech, which
encourage vastly different use-cases.  
For example, SUSAS was created to study speech under different stressful situations, such as operating a helicopter or being subject to psychiatric analysis. 
In contrast, MSP-Improv was recorded in a  
laboratory environment by student actors. Furthermore, the datasets are often labeled or validated by annotators who are 
distinct from the speakers. 
While SER is often framed as identifying speakers' true emotions, as would be necessary to support use-cases like call screening or mental health support, a more accurate characterization of datasets annotated this way would be
identifying how 
third parties perceive speakers' emotions.

\begin{table}
    \centering
    \scalebox{0.8}{
    \begin{tabular}{lclc}
\textcolor{violet}{Angry} & 76.14\% & \textcolor{violet}{Surprised} & 19.32\% \\
Neutral & 72.73\% & Boredom & 14.77\% \\
\textcolor{violet}{Sad} & 67.05\% & Joy & 9.09\% \\
\textcolor{violet}{Happy} & 65.91\% & Stressed & 7.95\% \\
\textcolor{violet}{Fearful} & 30.68\% & Calm & 6.82\% \\
\textcolor{violet}{Disgusted} & 27.27\% & \textcolor{teal}{Dominance} & 5.68\% \\
\textcolor{teal}{Valence/Sentiment} & 22.73\% & Frustrated & 3.41\% \\
\textcolor{teal}{Arousal/Activation} & 20.45\% & Excited & 2.27\% \\
\hline
Other/Unspecified & 21.59\% \\
    \end{tabular}
    }
    \caption{Percent of papers that use each emotion label, ordered by frequency. Color-coding is as \Cref{tab:datasets}.}
    \label{tab:emotion_labels}
\end{table}

Further, our analysis of the specific emotions studied in each paper reveals that papers use these datasets selectively (\Cref{tab:emotion_labels}). Papers study angry, neutral, sad, and happy twice as often as other emotions, even though datasets support other labels, and there is no clear indication that these emotions would better suit the motivations in \Cref{tab:motivations}.
IEMOCAP exemplifies this selective use. Almost all ($>90\%$) of papers that use IEMOCAP use categorical labels, but far fewer (2-20\%) use dimensional labels. Even within these categories usage is not consistent. While 19.44\% of IEMOCAP papers use Valence/Sentiment labels, only 2.78\% use Dominance.

\begin{figure}
    \centering
    \includegraphics[width=\linewidth]{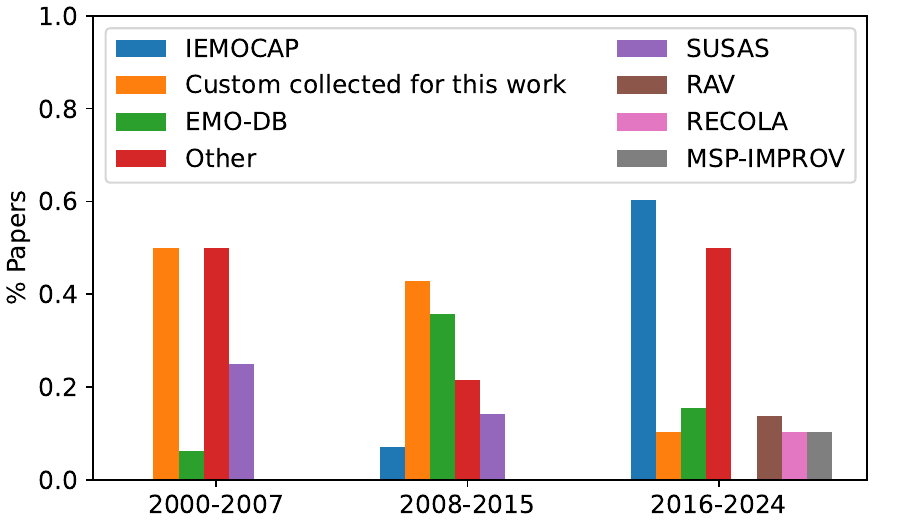}
    \caption{Percent of papers in each time window that use the specified dataset in their experiments.}
    \label{fig:datasets_overtime}
\end{figure}

Curiously, we find that dataset usage varies over time (\Cref{fig:datasets_overtime}). One notable trend is the decline in data that was custom-collected for the specific paper, which are sometimes, but not always, more directly constructed to support the motivating task. Although 22.73\% of the papers in our sample use custom-built datasets overall (\Cref{tab:datasets}), they are far more common in pre-2016 papers than 2016-2024 papers. 
Use of SUSAS has also declined, in contrast to IEMOCAP which accounts for almost 60\% all data used in 2016-2024.
The shifts in dataset usage are generally inconsistent with the shifts in motivations shown in \Cref{fig:motivations_overtime}. For instance, there is no corresponding rise in interest for a use case particularly suited to IEMOCAP that would justify its increased popularity in 2016-2024 as compared to 2008-2015 (IEMOCAP was released in 2008).

\paragraph{Do datasets align with stated motivations?}

\begin{figure}
    \centering
    \includegraphics[width=\linewidth]{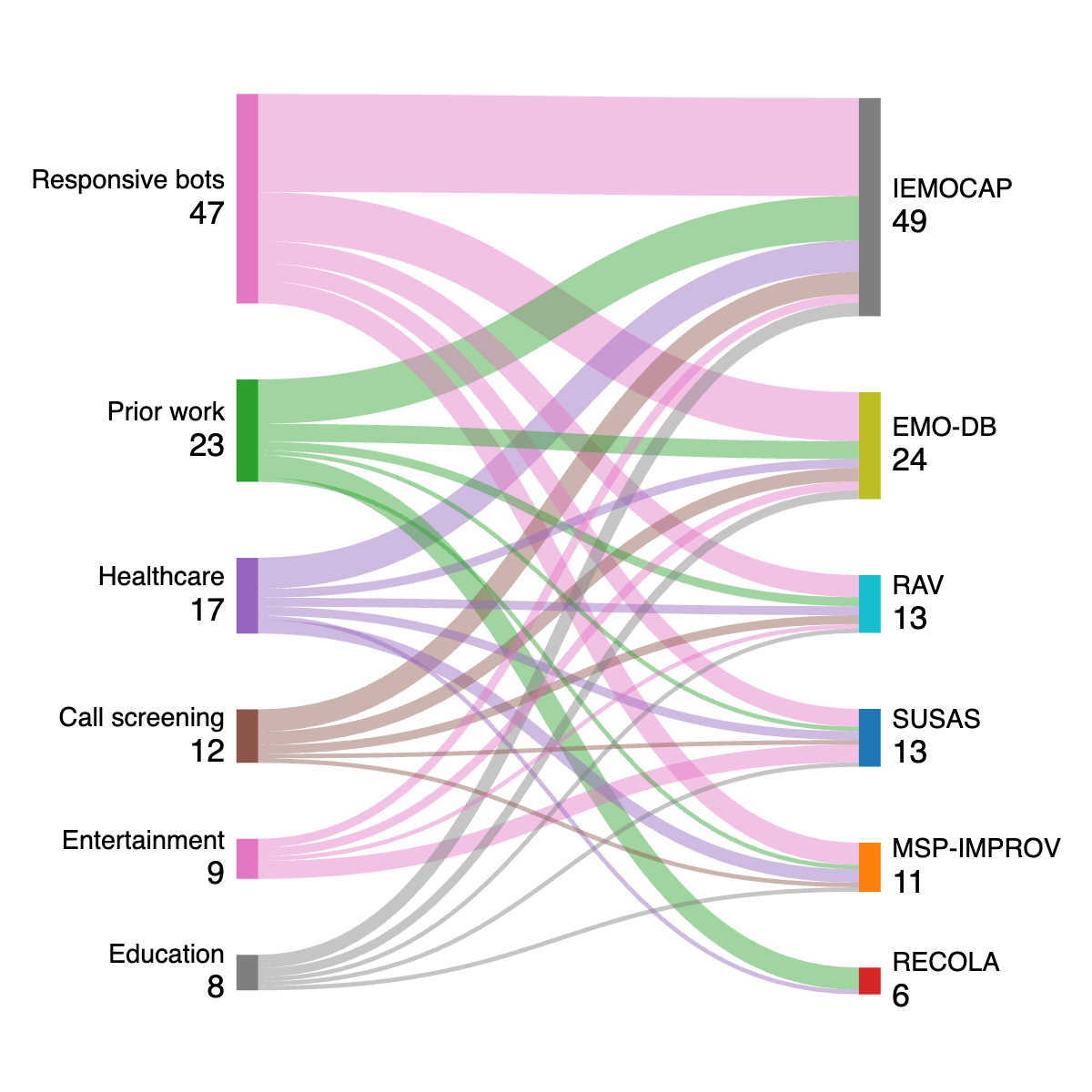}
    \caption{Mapping between most common stated motivations and datasets.}
    \label{fig:overview}
\end{figure}

In \Cref{fig:overview}, we display the mapping between the most common stated motivations and the six popular datasets, showing a notable lack of pattern. Despite the widely divergent downstream applications, at least one paper with every stated motivation used IEMOCAP. Similarly, although responsive bots or better human-computer interaction has been a persistent motivation for SER (\Cref{fig:motivations_overtime}), none of the popular data sets were constructed for this application (\Cref{tab:datasets}), yet they continue to be used for it (\Cref{fig:overview}). Datasets with acted emotions are often used for applications intended for spontaneous speech, yet research has shown that spontaneous emotional expressions differ in various ways from acted expressions \cite{juslin2018mirror}. 
Furthermore, most datasets are collected from spontaneous or acted human-human interactions but are frequently used to advance human-computer interaction, even though people express themselves differently and more subtly when interacting with chatbots \citep{kovacevic2024multimodal}, resulting in a significant domain mismatch.

Quantitative metrics of specific datasets and motivations yield similar mismatches. For example, 16 papers exclusively use IEMOCAP, but only 2 of them state Entertainment as a motivation, even though IEMOCAP's data collection protocol involving scripted and improvised acted speech is most similar to entertainment applications like movies and video games where speech is similarly acted. On the other hand, 16 papers listed Healthcare as a motivation, an application which requires uncovering a speakers true emotional state. While 3 of those papers exclusively use custom-collected data, and 2 papers evaluate over SUSAS (which was designed to capture simulated and actual stress), the remaining 11 papers all evaluate over almost entirely acted speech (e.g., 1 paper uses ASVP-ESD  \citep{landry2020asvp}, which is more spontaneous).

\section{Discussion}

While these results identify a mismatch between stated motivations and underlying datasets, research and deployment are not necessarily expected to be identical. Strong popularity of a small number of datasets, such as frequent use of IEMOCAP in recent years, potentially reflects increasing standardization in evaluation setups, e.g., using established benchmark datasets facilitates publication by enabling direct comparison with prior work. Standardization of tasks has typically been constructive for accelerating progress.
However, several characteristics of emotion prediction make the mismatch between stated motivations and empirical data problematic. We first discuss why this mismatch is potentially harmful and then offer avenues for reducing it.

\paragraph{Does the mismatch between data and motivations cause harm?}
First, there is substantial interest in deploying emotion recognition technology in decision-making contexts where misclassification can have harmful consequences. 
In their analysis of patent applications for uses of emotion recognition technology in workplace settings, \citet{boyd2023automated} find that many patent applications treat outputs of their technology as ground-truth representations of internal states that can inform consequential decisions like hiring, firing, and calling law enforcement. 
SER systems have been shown to generalize poorly across datasets, even ones collected in similar ways (e.g., acted speech in laboratory settings) \cite{Ri_2023}. 
There is also a fundamental difference between a person's true internal state and the emotions that an independent observer attributes to them, yet independent annotation to label or validate collected data is standard practice.
The implication that a system designed and evaluated over a particular dataset could be useful in a mismatched downstream application can lead to harmful errors. 

Additionally, people consider emotions private \cite{mcstay2020emotional,pyle2024us}, and even well-intentioned motivations are not necessarily well-received by people subject to this technology. 
In a survey of the attitudes of help seekers towards predictive tools for mental health risks, only about half of the respondents (49\%) were in favor of using risk prediction \cite{mantell2021attitudes}, and 
similar concerns about the use of AI emotion in education exist \cite{Stark2021}. 
Because of the sensitive nature of this topic,
researchers should set high standards for SER research and ensure there are clear downstream benefits 
rather than treating SER as a generic task to  
optimize for and assuming future adoption will be done responsibly.

\paragraph{What data are needed to support these motivations?}
\label{sec:discussion_motivation}

We next discuss how researchers could use task-appropriate datasets that reflect the intended downstream use cases, focusing for brevity  on the most commonly referenced motivation: Responsive bots.
Several studies collected datasets that are more aligned with this setup than the popular ones in \Cref{tab:datasets}. We identify a few examples out of the six papers that list responsive bots as a motivation and custom collect datasets. 
\citet{Wang2017} work with utterances from interactions with a spoken dialogue system, though emotion labels are separately crowd-sourced. \citet{Auberg2003WhyAH} propose a wizard-of-oz system for collecting this type of data, where a user believes they are interacting with a real computer system, but in reality a human is manipulating the system to create particular stimuli.
Conducting research with real human-system interactions from deployed systems can introduce privacy concerns, but collecting similar data is feasible in simulated laboratory settings where users can give informed consent. 

\paragraph{What use-cases can we support with this data?}
\label{sec:discussion_data}
Alternatively, researchers could better align motivations and datasets by pursuing motivations that popular SER datasets do support.
For example, because of its annotation setup, IEMOCAP offers a unique opportunity to investigate the alignment and discrepancies between the perception of emotions---i.e., the annotator's labels of emotions---and the intended portrayal of emotion---i.e., self-assessed emotions of the actors. Detection of emotions in acted speech could be used to investigate questions around emotional portrayals in acted media such as television shows \cite{zhou2024once}, which has been of interest in gender studies \cite{garcia_women_2016} and media theory \cite{Gorton_2009,garcia_emotions_2016} literature.

\paragraph{Conclusions}

Our analysis uncovers substantial gaps in how SER research is motivated and conducted. 
This mismatch exacerbates the risks associated with SER deployment, especially as the technology becomes increasingly commercialized. We recommend ways to reduce this gap, by shifting datasets to match motivations or motivations to match datasets. Given the sensitivity of emotion recognition technology, we urge researchers to critically evaluate the motivations behind SER projects, ensuring they are well-justified and that experiments meaningfully support them.
    
\section{Ethical Considerations and Limitations}

The primary limitation of our work is its reliance on a specific data sample. Although we carefully chose a range of search terms to identify SER papers and we stratify our data sample by year and publication venue, it is possible that analysis of a broader set of papers could yield different findings. 

\section{Bibliographical References}
\bibliographystyle{lrec2026-natbib}
\bibliography{lrec2026-example,citation}
\bibliographystylelanguageresource{lrec2026-natbib}
\bibliographylanguageresource{languageresource}

\appendix

\clearpage

\section{Examples of coded motivations}
\label{sec:app_examples}

\begin{minipage}{1.0\textwidth}
    \centering
    \scalebox{0.8}{
    \begin{tabular}{p{8cm}p{10cm}}
    \hline
Responsive bots: other HCI systems & ``These models could lead to computer agents and robots that more naturally and functionally blend into human society'' \citep{Provost2009EvaluatingEA} \\
\hline
Responsive bots: voice assistants; Responsive bots: other HCI systems & ``Speech emotion recognition is becoming more and more important for many applications related to human computer interactions, especially for spoken dialogue systems. With emotion recognition from users' speech, a better user experience can be achieved'' \citep{Wang2017} \\
\hline
Healthcare (mental health); Responsive bots: car voice assistants; Education; Other  & ``In affective computing field, emotion recognition from speech plays a very important role, and has received much attention over the past few decades... It has been proven useful in many applications which require human-machine interaction, e.g., in-car board system, diagnostic tool for therapists, automatic translation systems, computer tutorial applications'' \citep{Song2020FeatureSB} \\ \hline
Call screening & "In a call-center, such a system should be able to determine in a critical phase of the dialogue if the call should be passed over to a human operator." \citep{Huber2000RecognitionOE} \\ \hline
Paralinguistics / behavioral studies & ``Speech signals carry rich information on an individual’s emotional states, expressed through both paralinguistic and semantic cues.'' \citep{Oliveira2023LeveragingSI} \\ \hline
Social companion bots; Video games, toys, entertainment; Education; Other  & ``Research on emotion recognition can advance many applications, like distance education, social robots, video games, affective mirrors and many others.'' \citep{Li2023MusicTA} \\ \hline
Lie detection; Call screening; Healthcare (mental health & ``Especially in the field of human-machine interaction (HCI), growing interest can be observed in recent years. In addition, the detection of lies, monitoring of call centers and medical diagnoses are often claimed as promising application scenarios for speech emotion recognition.'' \citep{Mao2019DeepLO} \\
\hline
Prior work & ``While the majority of traditional research in emotional speech recognition has utilized a single database for analysis, it is becoming clear that the lack of sufficiently large databases with varied emotion types is a significant hindrance to the design of a robust emotion classification system.'' \citep{sun2012preliminary} \\
    \end{tabular}
    }
    \captionsetup{hypcap=false}
    \captionof{table}{Examples of human-selected text snippets and coded motivations, with at least one example per motivation. We remove citations from within quotations for readability.}
    \label{tab:data_examples}
\end{minipage}

\clearpage

\onecolumn
\footnotesize

{\setlength{\tabcolsep}{3pt} 
\begin{longtable}{%
  >{\raggedright\arraybackslash}p{0.16\textwidth}
  >{\raggedright\arraybackslash}p{0.12\textwidth}
  >{\raggedright\arraybackslash}p{0.2\textwidth}
  >{\raggedright\arraybackslash}p{0.2\textwidth}
  >{\raggedright\arraybackslash}p{0.22\textwidth}
}
\caption{All 88 manually coded papers}\label{tab:allpapers}\\
\toprule
 & Venue & Motivations & Datasets & Emotions \\
\midrule
\endfirsthead

\toprule
 & Venue & Motivations & Datasets & Emotions \\
\midrule
\endhead

\midrule
\multicolumn{5}{r}{\textit{Continued on next page}}\\
\endfoot

\bottomrule
\endlastfoot

\citet{Huber2000RecognitionOE} & Interspeech & Call screening & Custom & Angry, Other or Unspecified \\ \hline
\citet{Amir2001ClassifyingEI} & Interspeech & Prior work & Universidad Politecnica de Madrid & Angry, Sad, Happy, Neutral \\ \hline
\citet{Nogueiras2001SpeechER} & Interspeech & Healthcare, Entertainment & Interface Database & Surprised, Joy, Angry, Fearful, Disgusted, Sad, Neutral \\ \hline
\citet{Li2002RobustSR} & Interspeech & Prior work & Custom & Angry, Other or Unspecified, Sad, Neutral \\ \hline
\citet{Makarova2002RUSLANAAD} & Interspeech & Entertainment, Other & RUSLANA & Neutral, Surprised, Happy, Angry, Fearful, Sad \\ \hline
\citet{Rahurkar2003FrequencyDB} & Interspeech & Paralingusitic / behavioral studies, RB: voice assistants, Entertainment, RB: other HCI systems & SOQ & Neutral, Stressed \\ \hline
\citet{Auberg2003WhyAH} & Interspeech & RB: other HCI systems, RB: voice assistants & Custom & Other or Unspecified \\ \hline
\citet{Kwon2003EmotionRB} & Interspeech & Entertainment, RB: other HCI systems & SUSAS, FAU Aibo Emotion Corpus & Stressed, Angry, Boredom, Happy, Neutral, Sad \\ \hline
\citet{Kim2005IntegratingIF} & Interspeech & RB: other HCI systems & Custom & Arousal/Activation, Valence or sentiment \\ \hline
\citet{Shafran2005ACO} & ICASSP & RB: other HCI systems, Other, Call screening, Education & Custom & Valence or sentiment \\ \hline
\citet{Burkhardt2006DetectingAI} & Interspeech & Call screening & Custom & Angry \\ \hline
\citet{Gupta2007TwostreamER} & Interspeech & Call screening & EPST, Custom & Angry, Happy, Neutral \\ \hline
\citet{CasaleSalvatore2007MultistyleCO} & Speech Communication & RB: other HCI systems, Healthcare, Other, Entertainment, Education & SUSAS & Angry, Stressed, Neutral \\ \hline
\citet{Li2007StressAE} & ICASSP & Healthcare, Other, Lie detection, Entertainment & SUSAS, African Elephant Emotional Arousal Dataset, Custom & Other or Unspecified, Angry, Arousal/Activation,Neutral \\ \hline
\citet{Vlasenko2007CombiningFA} & Interspeech & Prior work & EMO-DB, SUSAS & Happy, Angry, Fearful, Boredom, Disgusted, Stressed, Neutral \\ \hline
\citet{Grimm2007SupportVR} & ICASSP & RB: other HCI systems & VAM & Valence or sentiment, Dominance, Other or Unspecified \\ \hline
\citet{Jones2008SpeechIW} & Interspeech & Call screening, Other, RB: car voice assistants, Entertainment, Lie detection & Custom & Happy, Boredom, Sad, Surprised, Angry, Other or Unspecified, Excited, Fearful, Frustrated \\ \hline
\citet{Vlasenko2008BalancingSC} & Interspeech & RB: other HCI systems & EMO-DB, SUSAS & Angry, Boredom, Disgusted, Fearful, Joy, Neutral, Sad, Stressed \\ \hline
\citet{Vlasenko2009ProcessingAS} & Interspeech & RB: other HCI systems & Kiel Corpus of Read Speech , EMO-DB, FAU Aibo Emotion Corpus & Angry, Other or Unspecified, Neutral, Joy \\ \hline
\citet{Provost2009EvaluatingEA} & Interspeech & RB: other HCI systems & IEMOCAP & Angry, Happy, Sad, Neutral, Valence or sentiment, Arousal/Activation \\ \hline
\citet{Yun2009SpeechER} & ICASSP & Call screening, Entertainment, RB: other HCI systems & EMO-DB & Angry, Fearful, Disgusted, Sad, Boredom, Neutral, Happy \\ \hline
\citet{LpezMoreno2009SpeakerDE} & Interspeech & RB: car voice assistants, Entertainment, Call screening & SUSAS & Neutral, Stressed \\ \hline
\citet{Giannakopoulos2009ADA} & ICASSP & Other & Custom & Valence or sentiment, Arousal \\ \hline
\citet{Sanchez2010DomainAA} & Interspeech & Call screening & Custom & Fearful, Sad, Neutral \\ \hline
\citet{Wu2011EmotionRO} & IEEE Transactions on Affective Computing & Healthcare, Education & Custom & Angry, Happy, Neutral, Sad \\ \hline
\citet{sun2012preliminary} & Interspeech & Prior work & EMO-DB, EPST, EMA & Neutral, Angry, Sad, Happy \\ \hline
\citet{Han2013ActiveLF} & Interspeech & Prior work & SEMAINE & Arousal/Activation, Dominance, Valence or sentiment, Other or Unspecified \\ \hline
\citet{Matsumiya2014DatadrivenGO} & Interspeech & Other & Custom & Angry, Sad, Happy, Fearful, Neutral \\ \hline
\citet{Meyer2014ANE} & Speech Communication & RB: other HCI systems & EMO-DB & Fearful, Disgusted, Happy, Boredom, Neutral, Sad, Angry \\ \hline
\citet{Gaka2015SystemSS} & Interspeech & Call screening & Custom & Angry, Stressed, Neutral \\ \hline
\citet{Zaidan2016MFCCGF} & Advances in Machine Learning and Signal Processing & RB: other HCI systems & EMO-DB & Happy, Sad, Fearful, Angry, Disgusted, Boredom, Neutral \\ \hline
\citet{Zhang2017InteractionAT} & Interspeech & RB: other HCI systems, Other, Entertainment, Education & IEMOCAP & Angry, Happy, Sad, Neutral, Other or Unspecified \\ \hline
\citet{Wang2017} & ICASSP & RB: other HCI systems, RB: voice assistants & Custom & Neutral, Happy, Sad, Angry \\ \hline
\citet{Bertero2017AFL} & ICASSP & Prior work & Custom & Angry, Happy, Sad \\ \hline
\citet{Rathner2018StateOM} & Interspeech & Healthcare & Custom & Arousal/Activation, Valence or sentiment \\ \hline
\citet{Yang2018PredictingAA} & Interspeech & Prior work & RECOLA, SEMAINE & Valence or sentiment, Arousal \\ \hline
\citet{Han2018TowardsCA} & ICASSP & Prior work & RECOLA & Arousal/Activation, Valence or sentiment \\ \hline
\citet{Zhao2018ExploringSR} & Interspeech & Prior work & CHEAVD, IEMOCAP & Angry, Disgusted, Happy, Sad, Surprised, Neutral, Other or Unspecified \\ \hline
\citet{Huang2018SpeechER} & Interspeech & Paralingusitic / behavioral studies & IEMOCAP & Angry, Happy: Happy + Excited, Neutral, Sad \\ \hline
\citet{Yenigalla2018SpeechER} & Interspeech & Prior work & IEMOCAP & Neutral, Happy, Sad, Angry \\ \hline
\citet{Milner2019ACS} & ASRU & Prior work & eNTERFACE, RAVDESS (RAV), IEMOCAP, MOSEI & Angry, Happy, Sad, Surprised, Disgusted, Fearful, Neutral, Frustrated, Excited, Calm, Other or Unspecified \\ \hline
\citet{Bao2019CycleGANBasedES} & Interspeech & RB: other HCI systems & IEMOCAP, MSP-IMPROV, TEDLIUM (release 2) & Angry, Happy: Happy + Excited, Sad, Neutral \\ \hline
\citet{Mao2019DeepLO} & Interspeech & Healthcare, Call screening, Lie detection & IEMOCAP, CASIA & Sad, Neutral, Angry, Fearful, Happy, Surprised \\ \hline
\citet{Oates2019RobustSE} & Interspeech & Prior work & EMO-DB, RECOLA, eNTERFACE, Polish Emotional Speech Database & Angry, Boredom, Fearful, Joy, Neutral, Sad, Disgusted, Happy, Surprised, Arousal/Activation, Valence or sentiment \\ \hline
\citet{Latif2019MultiTaskSA} & IEEE Transactions on Affective Computing & Healthcare, Call screening, RB: other HCI systems & IEMOCAP, MSP-IMPROV & Angry, Happy, Neutral, Sad \\ \hline
\citet{Yoon2019SpeechER} & ICASSP & RB: other HCI systems, Paralingusitic / behavioral studies & IEMOCAP & Happy: Happy + Excited, Angry, Neutral, Sad \\ \hline
\citet{Zhao2019AttentionEnhancedCT} & Interspeech & Prior work & IEMOCAP, FAU Aibo Emotion Corpus & Happy: Happy + Excited, Angry, Sad, Neutral, Emphatic, Positive, Rest \\ \hline
\citet{Triantafyllopoulos2019TowardsRS} & Interspeech & Prior work & RECOLA, EMO-DB, eNTERFACE, Mozilla Common Voice Audio Set & Arousal/Activation, Other or Unspecified \\ \hline
\citet{Sahu2019MultiModalLF} & Interspeech & Healthcare, Paralingusitic / behavioral studies, RB: voice assistants & IEMOCAP, MSP-IMPROV & Angry, Happy, Neutral, Sad \\ \hline
\citet{Paraskevopoulos2019UnsupervisedLR} & Interspeech & RB: other HCI systems & EMO-DB, IEMOCAP & Angry, Disgusted, Fearful, Joy, Boredom, Sad, Neutral, Happy \\ \hline
\citet{Huang2020DetectingUA} & IEEE Transactions on Affective Computing & Healthcare & Custom & Happy, Fearful, Angry, Surprised, Sad, Disgusted \\ \hline
\citet{Song2020FeatureSB} & IEEE Transactions on Affective Computing & Healthcare, Other, RB: voice assistants, Education & EMO-DB, eNTERFACE, FAU Aibo Emotion Corpus & Angry, Boredom, Disgusted, Fearful, Happy, Sad, Surprised \\ \hline
\citet{Shukla2020DoesVS} & IEEE Transactions on Affective Computing & Healthcare, Other & CREMA-D, RECOLA, RAVDESS (RAV), SEWA, IEMOCAP & Angry, Fearful, Disgusted, Happy, Sad, Neutral, Calm, Surprised, Arousal/Activation,Valence or sentiment \\ \hline
\citet{Li2020ADVISERAT} & ACL Demos & RB: voice assistants, Social Companion bots & IEMOCAP & Angry, Happy, Sad, Neutral, Arousal/Activation, Valence or sentiment \\ \hline
\citet{Latif2020DeepAE} & Interspeech & Prior work & IEMOCAP, MSP-IMPROV, DEMAND & Angry, Neutral, Sad, Happy: Happy + Excited \\ \hline
\citet{Chen2020AMF} & Interspeech & RB: other HCI systems & IEMOCAP & Happy: Happy + Excited, Neutral, Angry, Sad \\ \hline
\citet{Nediyanchath2020MultiHeadAF} & ICASSP & RB: voice assistants, Social Companion bots, RB: other HCI systems & IEMOCAP & Sad, Angry, Neutral, Happy \\ \hline
\citet{Sidorova2021TowardsDA} & IEEE Transactions on Affective Computing & Healthcare & Custom, Interface Database & Angry, Sad, Happy, Neutral \\ \hline
\citet{Huang2021AnAS} & Interspeech & Prior work & MSP-Podcasts & Neutral, Angry, Sad, Happy, Disgusted \\ \hline
\citet{Wu2021ANM} & IEEE Transactions on Affective Computing & Prior work & RECOLA, IEMOCAP & Arousal/Activation, Valence or sentiment \\ \hline
\citet{Tzirakis2021SpeechER} & ICASSP & RB: other HCI systems & SEWA & Arousal/Activation,Other or Unspecified, Valence or sentiment \\ \hline
\citet{Su2021ACC} & SLT & RB: other HCI systems, RB: car voice assistants & IEMOCAP, MSP-IMPROV, CIT: USC CreativeIT Corpus & Arousal/Activation, Valence or sentiment \\ \hline
\citet{Peng2021EfficientSE} & ICASSP & Call screening, Healthcare, RB: other HCI systems & IEMOCAP & Angry, Sad, Neutral, Happy: Happy + Excited \\ \hline
\citet{Li2022ExplorationOA} & SLT & Prior work & IEMOCAP, RAVDESS (RAV) & Angry, Happy: Happy + Excited, Neutral, Happy, Calm, Sad, Fearful, Surprised, Disgusted \\ \hline
\citet{Pham2022LearningCF} & IEEE Transactions on Affective Computing & RB: other HCI systems & RAVDESS (RAV), SAVEE, VidTIMIT, GRID, GEMEP & Neutral, Calm, Happy, Sad, Angry, Fearful, Surprised, Disgusted \\ \hline
\citet{Prabhu2022EndtoEndLU} & IEEE Transactions on Affective Computing & Prior work & AVEC’16, MSP-Conversation & Valence or sentiment, Arousal \\ \hline
\citet{Hu2022MultipleET} & Interspeech & RB: other HCI systems & IEMOCAP & Neutral, Angry, Happy, Sad \\ \hline
\citet{Kang2022SpeechEQSE} & Interspeech & RB: other HCI systems & Custom & Other or Unspecified \\ \hline
\citet{Feng2022SemiFedSERSL} & Interspeech & Other, Healthcare, RB: voice assistants, Education & IEMOCAP, MSP-IMPROV & Happy, Neutral, Angry, Other or Unspecified \\ \hline
\citet{Ye2022GMTCNetGM} & Speech Communication & RB: other HCI systems & CASIA, EMO-DB, RAVDESS (RAV), SAVEE & Angry, Boredom, Calm, Disgusted, Fearful, Happy, Neutral, Sad, Surprised \\ \hline
\citet{Hou2022MultiViewSE} & IEEE/ACM Transactions on Audio, Speech, and Language Processing  & RB: other HCI systems & IEMOCAP, EMO-DB & Angry, Sad, Neutral, Boredom, Fearful, Disgusted, Happy: Happy + Excited \\ \hline
\citet{Goncalves2022ImprovingSE} & Interspeech & Prior work & CREMA-D, MSP-Face & Happy, Fearful, Disgusted, Angry, Sad, Neutral \\ \hline
\citet{Li2023MusicTA} & IEEE/ACM Transactions on Audio, Speech, and Language Processing  & Other, Social Companion bots, Entertainment, Education & IEMOCAP & Neutral, Happy: Happy + Excited, Sad, Angry, Valence or sentiment, Arousal/Activation,Dominance \\ \hline
\citet{Chien2023AchievingFS} & ICASSP & RB: other HCI systems & IEMOCAP & Neutral, Happy, Angry, Frustrated, Sad \\ \hline
\citet{Meng2023WhatIL} & Interspeech & Prior work & CREMA-D & Other or Unspecified \\ \hline
\citet{Cahyawijaya2023CrossLingualCA} & Interspeech & Prior work & CREMA-D, ElderReact, ESD: Emotional Speech Database, TESS, YueMotion, CSED, IEMOCAP & Happy, Sad, Neutral, Disgusted, Fearful, Angry \\ \hline
\citet{Wu2023IntegratingER} & Interspeech & Prior work & IEMOCAP & Happy, Sad, Angry, Neutral \\ \hline
\citet{Xie2023FusionbasedSE} & Speech Communication & RB: other HCI systems, Other, Call screening, Entertainment, RB: car voice assistants, Education & RAVDESS (RAV), EMO-DB, SAVEE, EMOVO & Angry, Boredom, Fearful, Disgusted, Joy, Sad, Neutral, Happy, Surprised \\ \hline
\citet{Chen2023DSTDS} & ICASSP & Paralingusitic / behavioral studies & IEMOCAP, MELD & Happy: Happy + Excited, Angry, Sad, Neutral, Other or Unspecified \\ \hline
\citet{Oliveira2023LeveragingSI} & Interspeech & Paralingusitic / behavioral studies & MSP-Podcasts & Dominance, Arousal/Activation,Valence or sentiment \\ \hline
\citet{Bansal2023OnTE} & Interspeech & RB: other HCI systems, RB: car voice assistants, Call screening & IEMOCAP & Angry, Happy, Sad, Neutral \\ \hline
\citet{Mitra2023InvestigatingSR} & ICASSP & RB: voice assistants, RB: other HCI systems, Healthcare & MSP-Podcasts & Dominance, Valence or sentiment \\ \hline
\citet{Li2023ASRAE} & Interspeech & Prior work, RB: car voice assistants & IEMOCAP, MELD, CMU-MOSI & Angry, Happy: Happy + Excited, Neutral, Valence or sentiment, Disgusted, Fearful, Sad, Surprised, Joy \\ \hline
\citet{Zhang2023ADA} & Interspeech & RB: voice assistants, Social Companion bots & IEMOCAP, MELD & Happy: Happy + Excited, Angry, Neutral, Sad, Joy, Disgusted, Fearful, Surprised \\ \hline
\citet{Gao2023TwostageFO} & Interspeech & RB: other HCI systems & IEMOCAP & Angry, Neutral, Sad, Happy: Happy + Excited \\ \hline
\citet{Lashkarashvili2024ParameterEF} & ICASSP & RB: other HCI systems & IEMOCAP & Happy: Happy + Excited, Sad, Angry, Neutral, Other or Unspecified \\ \hline
\citet{Goron2024ImprovingDG} & ICASSP & RB: other HCI systems, Healthcare, Call screening & ASVP-ESD, IEMOCAP, RAVDESS (RAV), CREMA-D, CAFE, EMO-DB & Happy: Happy + Excited, Happy, Sad, Angry, Neutral \\ \hline
\citet{Wang2024GradientBasedDR} & ICASSP & RB: other HCI systems, RB: voice assistants & RAVDESS (RAV), SAVEE, TESS & Calm, Happy, Sad, Angry, Fearful, Surprised, Disgusted, Neutral \\ \hline
\end{longtable}

\normalsize
\twocolumn
\clearpage

\normalsize
\twocolumn
\clearpage

\end{document}